%% file: paper_6096.tex
\documentclass[runningheads]{llncs}
\usepackage[T1]{fontenc}
\usepackage{graphicx,verbatim}
\usepackage{amsmath}
\usepackage{amssymb}
\usepackage{multirow}

\usepackage{booktabs}

\usepackage[dvipsnames]{xcolor} 

\begin{document}
%
\title{Unrolled Reconstruction with Integrated Super-Resolution for Accelerated 3D LGE MRI}

\titlerunning{Unrolled Reconstruction with Integrated Super-Resolution}
%
\author{
Md Hasibul Husain Hisham\inst{1} \and
Shireen Elhabian\inst{1} \and
Ganesh Adluru\inst{2} \and
Jason Mendes\inst{2} \and
Andrew Arai\inst{2,3} \and
Eugene Kholmovski\inst{2} \and
Ravi Ranjan\inst{3} \and
Edward DiBella\inst{2}
}

\authorrunning{M. H. H. Hisham et al.}

\institute{
Kahlert School of Computing, University of Utah, Salt Lake City, UT, USA \and
Radiology \& Imaging Sciences, University of Utah, Salt Lake City, UT, USA \and
Cardiology, University of Utah, Salt Lake City, UT, USA
}

  
\maketitle              
\begin{abstract}
Accelerated 3D late gadolinium enhancement (LGE) MRI requires robust reconstruction methods to recover thin atrial structures from undersampled k-space data. While unrolled model-based networks effectively integrate physics-driven data consistency with learned priors, they operate at the acquired resolution and may fail to fully recover high-frequency detail. We propose a hybrid unrolled reconstruction framework in which an Enhanced Deep Super-Resolution (EDSR) network replaces the proximal operator within each iteration of the optimization loop, enabling joint super-resolution enhancement and data consistency enforcement. The model is trained end-to-end on retrospectively undersampled preclinical 3D LGE datasets and compared against compressed sensing, Model-Based Deep Learning (MoDL), and self-guided Deep Image Prior (DIP) baselines. Across acceleration factors, the proposed method consistently improves PSNR and SSIM over standard unrolled reconstruction and better preserves fine cardiac structures, leading to improved LA (left atrium) segmentation performance. These results demonstrate that integrating super-resolution priors directly within model-based reconstruction provides measurable gains in accelerated 3D LGE MRI.


\end{abstract}

\input{sections/introduction}
\input{sections/related_works}
\input{sections/dataset}
\input{sections/methods}
\input{sections/results}

\input{sections/conclusion}

\bibliographystyle{splncs04}
\bibliography{ref-6096}
%




\end{document}

%% file: sections/introduction.tex
\section{Introduction}

Late gadolinium enhancement (LGE) MRI is clinically useful for visualizing left atrial fibrosis in patients with atrial fibrillation, enabling pre-ablation planning and post-procedure assessment~\cite{mcgann2008new}. Current 3D LGE-MRI protocols, while effective, require lengthy acquisition times (typically 7--20 minutes), resulting in patient discomfort and increased susceptibility to motion artifacts~\cite{kecskemeti2013volumetric}. Accelerating acquisition via k-space undersampling is therefore a critical clinical need. But recovering diagnostic-quality images from highly undersampled measurements poses a challenging ill-posed inverse problem, particularly for 3D LGE imaging where precise visualization of thin atrial walls is essential. In this setting, even subtle loss of high-frequency detail can obscure fibrosis boundaries and compromise downstream quantitative analysis, making reconstruction fidelity at fine spatial scales especially critical.

Traditional compressed sensing (CS) reconstruction enables acceleration by exploiting the sparsity of MRI data, formulating reconstruction as a regularized inverse problem that balances data fidelity with handcrafted priors \cite{feng2013highly,jin2016general,adluruisotropic,tv1}. 
CS-based reconstruction has since become the clinical standard for accelerated MRI acquisition~\cite{jaspan2015compressed}.
However, under higher acceleration factors CS may attenuate fine structural detail due to the smoothing effects of sparsity-promoting regularization. Moreover, CS requires careful hand-tuning of regularization parameters and remains computationally intensive, often requiring hours per volume. 

Supervised deep learning methods have demonstrated strong reconstruction performance~\cite{lu2020pfista,qin2018convolutional}. 
However, purely data-driven approaches may implicitly learn acquisition-specific biases and typically require large paired datasets of undersampled and fully-sampled acquisitions, which presents a fundamental barrier in cardiac imaging. In the cardiac domain, acquiring motion-free, fully sampled 3D LGE volumes at high spatial resolution is particularly challenging, limiting the availability of high-quality reference data.
Unsupervised methods offer a compelling alternative by eliminating this data dependency~\cite{moreno2021evaluation}. In particular, Deep Image Prior (DIP)-based approaches leverage the implicit bias of convolutional network architectures as a learned prior \cite{bell2023robust,hisham2025family}.
Nevertheless, such methods rely primarily on implicit architectural regularization and optimize network parameters independently for each image, without learning a shared prior across subjects. This can limit reconstruction robustness and stability under high acceleration, particularly when recovering fine anatomical detail.

Model-based unrolled reconstruction methods offer a principled middle ground, integrating learned image priors with physics-based data consistency through alternating proximal gradient and conjugate gradient steps \cite{aggarwal2018modl,liang2019deep}. Unlike purely data-driven approaches, unrolled methods explicitly incorporate the MRI forward model (SENSE encoding), promoting reconstructions that remain faithful to acquired measurements. 
However, in most implementations the learned proximal operator is designed primarily for denoising at the acquired spatial resolution, which may limit recovery of high-frequency anatomical detail under aggressive undersampling.
Enhanced Deep Super-Resolution (EDSR) networks \cite{lim2017enhanced,zhang2018residual} have demonstrated strong performance in recovering high-frequency image detail.
This raises a key question: \textit{can super-resolution priors be embedded directly within the unrolled reconstruction framework to enhance fine structural recovery while preserving strict data consistency?} Rather than applying super-resolution as a post-processing step, integrating it within each unrolled iteration may enable joint optimization of resolution enhancement and measurement fidelity, potentially yielding improvements beyond what either approach achieves independently.

In this work, we investigate this hypothesis on preclinical (canine) 3D LGE cardiac MRI data. 
Specifically, we integrate an Enhanced Deep Super-Resolution (EDSR) network within each iteration of a model-based unrolled reconstruction framework, replacing the conventional denoising proximal operator with a super-resolution–aware prior.

Super-resolution for MRI image restoration has been explored through CNN-based methods \cite{kobayashi2020improving,qiu2021gradual,pham2019multiscale,du2020super}, Low-rank based methods \cite{shi2015lrtv,cherukuri2019deep}, SMORE method~\cite{zhao2019applications}, etc. 
However, these methods have been primarily validated on brain or musculoskeletal imaging, leaving cardiac applications largely unexplored. The cardiac domain presents unique challenges: respiratory and cardiac motion introduce artifacts that compound with undersampling degradation, and thin structures such as the left atrial wall demand high fidelity in atrial fibrosis settings. This motivates integrating super-resolution capability directly within the reconstruction loop.

We compare the proposed approach against traditional CS~\cite{adluruisotropic}, unrolled Model-Based Deep Learning (MoDL) with a U-Net denoiser~\cite{aggarwal2018modl}, and self-guided Deep Image Prior (DIP) as an unsupervised reference~\cite{hisham2025family}.

For MRI, downstream tasks such as quality assessment~\cite{sultan2024hamil}, classification~\cite{wahlang2022brain}, and segmentation~\cite{morshuis2024segmentation} are typically studied on fully-sampled images. Evaluating them on reconstructed outputs provides a more clinically grounded assessment beyond PSNR/SSIM alone.
We additionally evaluate downstream LA (left atrium) segmentation, reflecting the clinical sensitivity of fibrosis delineation to fine structural preservation.

Our contributions are as follows:
\begin{itemize}
    \item We propose a super-resolution–enhanced unrolled reconstruction framework that embeds EDSR directly within the iterative optimization loop, enabling joint enhancement of high-frequency anatomical detail and physics-based data consistency for accelerated 3D LGE-MRI.
    \item We demonstrate that this integration yields consistent improvements in PSNR and SSIM over standard unrolled reconstruction across acceleration factors, and improves downstream LA segmentation performance, providing a more clinically grounded evaluation than reconstruction metrics alone.
\end{itemize}

%% file: sections/methods.tex
\section{Methodology}

\begin{figure}[t]
    \centering
    \includegraphics[width=\textwidth]{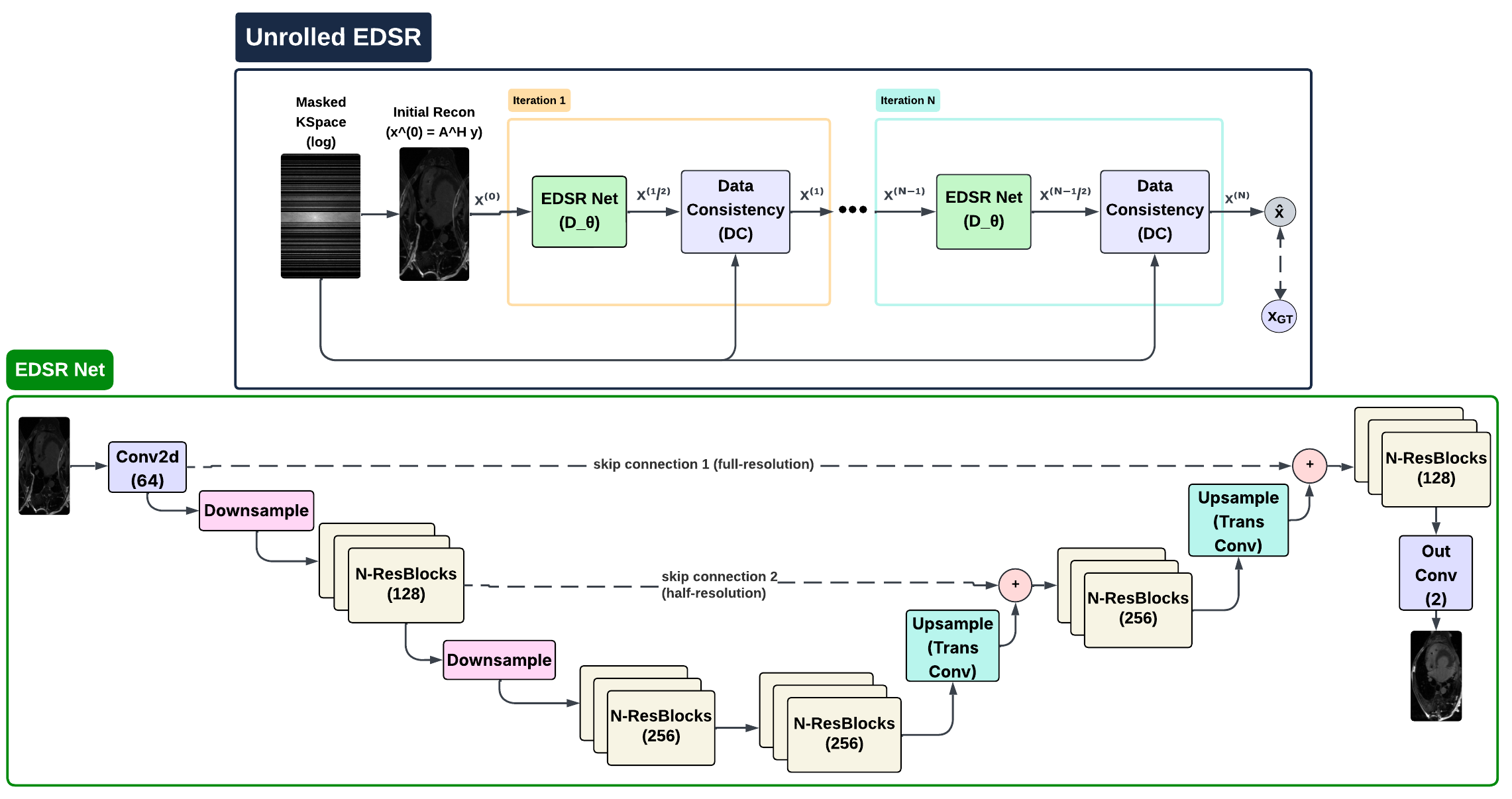}
    \caption{Proposed Unrolled EDSR framework. The EDSR Net serves as the learned proximal operator at each of the $N$ unrolled iterations, with weights $\phi$ shared across iterations. The initial reconstruction $\mathbf{x}^{(0)} = \mathbf{A}^H\mathbf{y}$ is the zero-filled adjoint, and data consistency (DC) enforces k-space fidelity via conjugate gradient at each step.
    }
    \label{fig:model}
\end{figure}

We address the problem of reconstructing a 2D MRI image from undersampled multi-coil 3D k-space measurements. For each axial slice, the input is multi-coil k-space data $\mathbf{y} \in \mathbb{C}^{N_c \times N_x \times N_y}$, where $N_c$ is the number of receiver coils and $N_x, N_y$ are the k-space dimensions. The acquisition is governed by a pseudo-random undersampling mask $\mathbf{M} \in \{0,1\}^{N_x \times N_y}$ with a densely sampled center, and coil sensitivity maps $\mathbf{S} \in \mathbb{C}^{N_c \times N_x \times N_y}$. The SENSE forward model per coil $c$ is:
\begin{equation}
    \mathbf{y}_c = \mathbf{M} \odot \mathcal{F}(\mathbf{S}_c \cdot \mathbf{x})
\end{equation}
where $\mathcal{F}$ denotes the 2D Fourier transform and $\odot$ is the element-wise product. The goal is to recover the reconstructed magnitude image $\hat{\mathbf{x}} \in \mathbb{R}^{N_x \times N_y}$ per slice. All methods described below operate slice-by-slice in 2D. The SENSE encoding operator is defined compactly as $\mathbf{A} = \mathbf{M} \odot \mathcal{F}\mathbf{S}$, with $\mathbf{A}^H$ denoting its adjoint.

\subsection{Baselines}

\noindent\textbf{Compressed Sensing.} The CS baseline solves the following regularized optimization problem per slice:
\begin{equation}
    \hat{\mathbf{x}} = \arg\min_{\mathbf{x}} \; \lambda_1 \|\mathbf{A}\mathbf{x} - \mathbf{y}\|_2^2 + \lambda_2 \text{TV}(\mathbf{x}) + \lambda_3 \text{BM3D}(\mathbf{x})
\end{equation}
where $\mathbf{y}$ is the acquired k-space data, $\text{TV}(\mathbf{x})$ enforces total variation regularization, and $\text{BM3D}(\mathbf{x})$ applies block-matching and 3D filtering to exploit non-local self-similarity~\cite{adluruisotropic}. The weights $\lambda_1$, $\lambda_2$, $\lambda_3$ are tuned via grid search. 

\noindent\textbf{Self-Guided DIP.} As an unsupervised baseline, we include self-guided Deep Image Prior~\cite{hisham2025family}, which requires no paired training data and optimizes network parameters independently per slice. The optimization objective is:
\begin{equation}
    \min_{\theta,\,\mathbf{z}} \; \sum_{c=1}^{N_c} \left\|\mathbf{A}_c \,\mathbb{E}_\eta[f_\theta(\mathbf{z}+\eta)] - \mathbf{y}_c\right\|_2^2 + \alpha\left\|\mathbb{E}_\eta[f_\theta(\mathbf{z}+\eta)] - \mathbf{z}\right\|_2^2
\end{equation}
where $f_\theta$ is an untrained ResUNet, $\eta$ represents random Gaussian perturbations, and $\alpha$ controls the strength of self-regularization as an implicit denoising mechanism. The final reconstruction is:
\begin{equation}
    \hat{\mathbf{x}} = \mathbb{E}_{\eta \sim \mathcal{P}_\eta}\!\left[f_{\theta^*}(\mathbf{z}^* + \eta)\right]
\end{equation}

\subsection{Proposed Method: Unrolled Reconstruction with EDSR}

The central contribution of this work is an unrolled model-based reconstruction framework in which the proximal operator at each iteration is instantiated as an EDSR network~\cite{lim2017enhanced,zhang2018residual,zamir2021multi}, replacing the conventional U-Net denoiser with a architecture that jointly performs image refinement and feature enhancement within the physics-constrained optimization loop. This design embeds learned image enhancement directly into each unrolled iteration, rather than treating it as a post-processing step, allowing data consistency to supervise the EDSR at every stage of reconstruction.

\noindent\textbf{Initialization.} The reconstruction is initialized from the zero-filled adjoint reconstruction:
\begin{equation}
    \mathbf{x}^{(0)} = \mathbf{A}^H \mathbf{y}
\end{equation}

\noindent\textbf{Denoiser step.} At each iteration $n$, the image estimate is refined via the EDSR proximal operator $\mathcal{E}_\phi$:
\begin{equation}
    \mathbf{x}^{(n+\frac{1}{2})} = \mathcal{E}_\phi\!\left(\mathbf{x}^{(n)}\right)
\end{equation}
As illustrated in Fig.~\ref{fig:model}, $\mathcal{E}_\phi$ follows a U-Net-style encoder-decoder with residual blocks, two downsampling stages, transposed convolution upsamplers, and full- and half-resolution skip connections, projecting to 2 output channels (real and imaginary). Parameters $\phi$ are shared across all $N$ iterations.

\noindent\textbf{Data consistency step.} The refined estimate is projected back onto the acquisition manifold via conjugate gradient optimization:
\begin{equation}
    \mathbf{x}^{(n+1)} = \arg\min_{\mathbf{x}} \; \|\mathbf{A}\mathbf{x} - \mathbf{y}\|_2^2 + \lambda\|\mathbf{x} - \mathbf{x}^{(n+\frac{1}{2})}\|_2^2
\end{equation}
with closed-form CG solution:
\begin{equation}
    \mathbf{x}^{(n+1)} = \left(\mathbf{A}^H\mathbf{A} + \lambda\mathbf{I}\right)^{-1}\!\left(\mathbf{A}^H\mathbf{y} + \lambda\mathbf{x}^{(n+\frac{1}{2})}\right)
\end{equation}
The final reconstruction is $\hat{\mathbf{x}} = \mathbf{x}^{(N)}$. The full pipeline is trained jointly end-to-end by minimizing the $\ell_2$ reconstruction loss against the fully-sampled ground truth:
\begin{equation}
    \mathcal{L} = \|\hat{\mathbf{x}} - \mathbf{x}_{GT}\|_2
\end{equation}

\subsubsection{Vanilla Unrolled Variant.}
For ablation purposes, we also evaluate a vanilla unrolled variant in which the EDSR proximal operator $\mathcal{E}_\phi$ is replaced by a standard U-Net denoiser $\mathcal{D}_\theta$, with all other components: initialization, data consistency, weight sharing, and training loss — held identical. This isolates the contribution of the EDSR architecture within the unrolled loop.

%% file: sections/results.tex
\section{Results}

\subsection{Implementation Details}

\subsubsection{Dataset}
We evaluate all methods on a preclinical 3D LGE cardiac MRI dataset of 24 subjects, acquired on a Siemens 3T scanner. Sequence parameters included TR = 2.7 ms, TE = 1.5 ms, with individually optimized inversion time (TI) determined via a TI scout sequence for optimal myocardial nulling. The acquisition achieves anisotropic resolution of 1.25$\times$1.25$\times$2.5 mm with approximately 50--60 axial slices per subject. Due to the higher heart rate in preclinical subjects, fewer readout lines are acquired per heartbeat compared to clinical protocols. A continuous low-dose gadolinium maintenance infusion is used to stabilize blood pool contrast throughout the scan, keeping TI constant. Motion compensation is achieved via ECG-gating targeting the diastolic phase and a one-dimensional respiratory navigator.

We obtained fully-sampled acquisitions for all the 24 preclinical subjects, enabling retrospective undersampling. Random undersampling masks with a densely sampled center region are applied retrospectively at acceleration factors $R \in \{4, 6\}$. Although the data is acquired as a 3D volume, we adopt a 2D slice-wise processing strategy to increase the number of available training samples and reduce GPU memory requirements.



\subsubsection{Training}
All experiments were implemented in PyTorch and trained on a single NVIDIA A100 40GB GPU. The 24 preclinical subjects were split 16/3/5 for training, validation, and testing, with all 3D volumes processed slice-by-slice in 2D. Coil sensitivity maps were estimated using Walsh method~\cite{griswold2006autocalibrated}. The pipeline is trained end-to-end with an $\ell_2$ loss using the Adam optimizer (lr $= 3\times10^{-5}$, with Cosine scheduler) for $1000$ epochs with batch size $32$. Data augmentation includes doing R=4 and R=6 undersampling on the slices, with 6\% of the Ky lines sampled in the center. Network weights $\phi$ are shared across all $N$ unrolled iterations, with $N=7$ selected by ablation. 

\subsection{Analysis}

\begin{figure}[t]
    \centering
    \includegraphics[width=\textwidth]{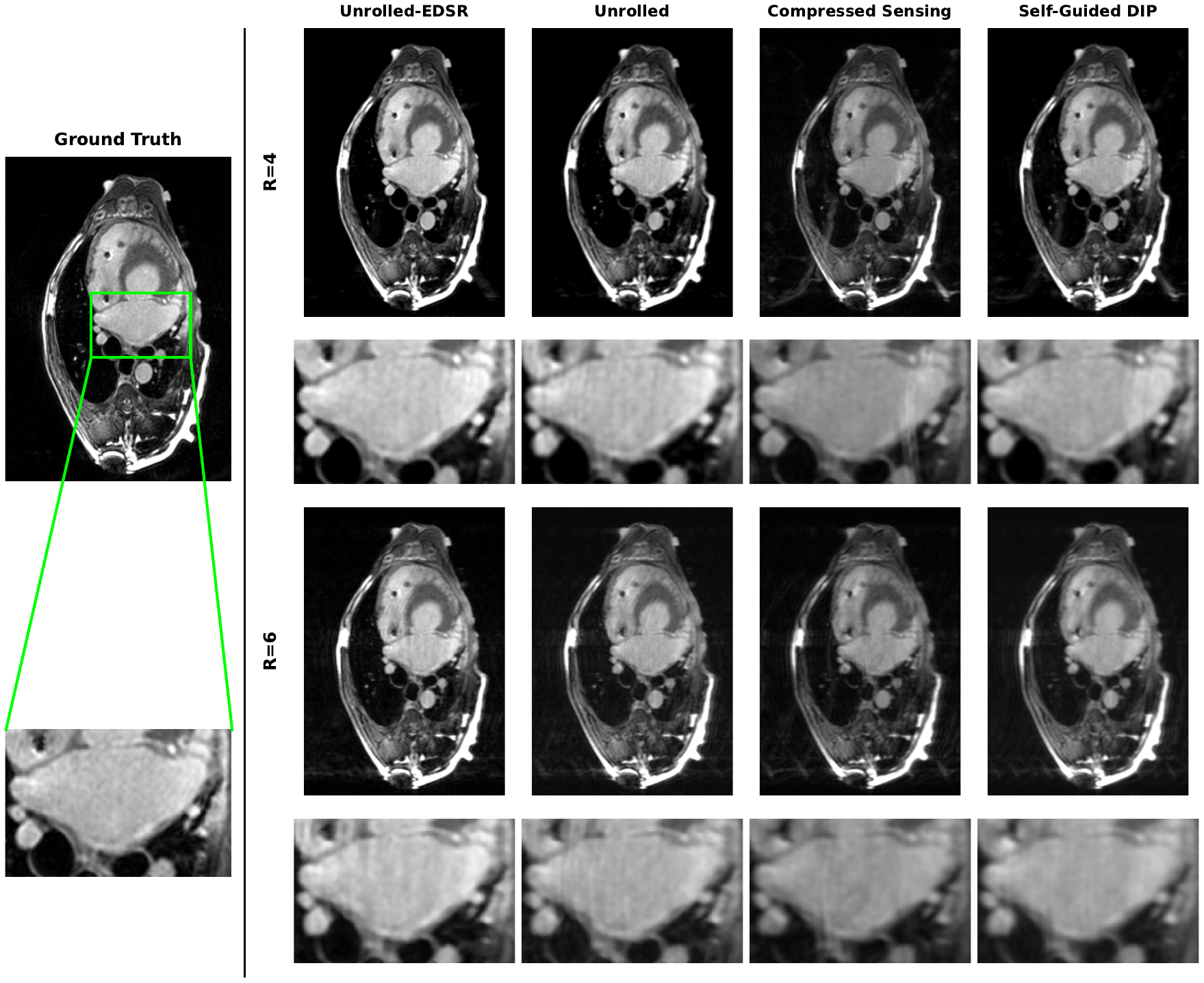}
    \caption{Qualitative comparison of reconstruction methods at acceleration factors $R=4$ and $R=6$. Each column shows a representative axial slice from a preclinical subject. At $R=4$, Unrolled+EDSR and Unrolled both recover fine cardiac structures faithfully, while Compressed Sensing introduces visible artifacts and DIP shows smoothing. At $R=6$, quality degrades across all methods, with Unrolled+EDSR best preserving structural detail.
    The LA is zoomed in for better visibility.
    }
    \label{fig:results}
\end{figure}

\subsubsection{Baselines}

Figure~\ref{fig:results} presents qualitative reconstructions across all methods at $R \in \{4, 6\}$. Table~\ref{tab:results} reports quantitative PSNR and SSIM metrics averaged across the 5 test subjects and both acceleration factors, corroborating the qualitative findings.

\begin{table}[]
\centering
\caption{Quantitative comparison of reconstruction methods averaged across 5 preclinical subjects at acceleration factors $R=4$ and $R=6$. Best results in \textbf{bold}.}
\label{tab:results}
\begin{tabular}{|l|cc|cc|}
\hline
\multirow{2}{*}{\textbf{Method}} & \multicolumn{2}{c|}{$R=4$} & \multicolumn{2}{c|}{$R=6$} \\
 & PSNR (dB) & SSIM & PSNR (dB) & SSIM \\
\hline
Compressed Sensing                & $31.6 \pm 1.0$ & $0.865 \pm 0.019$ & $28.6 \pm 1.4$ & $0.824 \pm 0.020$ \\
Self-Guided DIP            & $32.8 \pm 1.2$ & $0.894 \pm 0.018$ & $29.3 \pm 1.1$ & $0.855 \pm 0.015$ \\
Unrolled          & $35.1 \pm 1.2$ & $0.906 \pm 0.015$ & $32.4 \pm 1.5$ & $0.878 \pm 0.012$ \\
EDSR-Unrolled     & $\mathbf{35.6 \pm 1.1}$ & $\mathbf{0.915 \pm 0.014}$ & $\mathbf{32.9 \pm 1.6}$ & $\mathbf{0.889 \pm 0.015}$ \\
\hline
\end{tabular}
\end{table}

\subsubsection{Downstream Segmentation}

We assess LA segmentation performance on the reconstructed test slices to evaluate the clinical relevance of reconstruction quality. A U-Net segmentation model~\cite{vittikop2019automatic} was trained on the 19 training subjects (same subjects as the reconstruction task) using the fully-sampled ground truth segmentations, achieving a DSC (Dice Similarity Coefficient) of $0.928$ on the 5 held-out test subjects. This trained model was then applied directly to the reconstructed slices of the same 5 test subjects across all methods. Table \ref{tab:segmentation} shows the DSC evaluation across different methods.

\begin{table}[]
\centering
\caption{LA segmentation DSC on reconstructed test slices. Results at R = 4.}
\label{tab:segmentation}
\setlength{\tabcolsep}{10pt}
\begin{tabular}{|l|c|}
\hline
Method & DSC \\
\hline
Compressed Sensing        & $0.809$ \\
Self-Guided DIP           & $0.835$ \\
Unrolled           & $0.884$ \\
Unrolled EDSR (Ours)      & $\mathbf{0.893}$ \\
\hline
Ground Truth (upper bound) & $0.928$ \\
\hline
\end{tabular}
\end{table}

\subsubsection{Ablation Experiments.}

We ablate the number of unrolling iterations $N \in \{5, 7, 9, 11\}$ in the proposed Unrolled+EDSR framework at $R=4$, examining the trade-off between reconstruction quality and computational memory.

\begin{table}[]
\centering
\caption{Ablation on number of unrolling iterations $N$. Results at $R=4$.}
\label{tab:ablation}
\begin{tabular}{|c|cc|c|}
\hline
$N$ & PSNR (dB) & SSIM & GPU Memory (GB) \\
\hline
5  & $34.2 \pm 1.1$ & $0.897 \pm 0.012$ & 16.3 \\
7  & $35.6 \pm 1.1$ & $0.915 \pm 0.012$ & 17.4 \\
9  & $35.9 \pm 1.1$ & $0.923 \pm 0.012$ & 21.6 \\
11 & $35.9 \pm 1.1$ & $0.929 \pm 0.012$ & 27.6 \\
\hline
\end{tabular}
\end{table}

Table \ref{tab:ablation} shows that the reconstruction quality improves consistently from $N=5$ to $N=9$, with PSNR gains of 1.7 dB and SSIM gains of 0.026. Beyond $N=9$, performance plateaus — $N=11$ yields identical PSNR with only marginal SSIM improvement — while GPU memory increases by 28\% (21.6 to 27.6 GB). We therefore adopt $N=7$ as our default configuration, balancing strong reconstruction quality with practical memory requirements.

%% file: sections/conclusion.tex
\section{Conclusion}

This work demonstrated that embedding a super-resolution aware proximal operator within an unrolled model-based framework yields measurable gains in accelerated 3D LGE-MRI reconstruction, both in pixel-level fidelity and downstream LA segmentation performance.
Evaluated on the preclinical subjects at acceleration factors $R \in \{4, 6\}$, the proposed Unrolled+EDSR consistently outperforms standard unrolled reconstruction, compressed sensing, and self-guided DIP across both PSNR/SSIM and downstream left atrium segmentation. The segmentation results are particularly notable: the DSC gain of Unrolled+EDSR over Unrolled (0.893 vs.\ 0.884) suggests that the super-resolution prior recovers fine atrial detail beyond what numerical metrics alone capture.

Several limitations remain. First, self-guided DIP's lower performance relative to supervised methods is consistent with its known spectral bias~\cite{shi2022measuring}. This bias essentially limits recovery of high-frequency structures under aggressive undersampling. Second, evaluation is currently limited to preclinical data. Validation on clinical human subjects with respiratory motion and variable heart rates is needed. Future work will explore training losses to further align reconstruction quality with downstream clinical utility, and extend the framework to fully 3D unrolled optimization.
Moreover, we would like to investigate its applicability to 3D Dixon fat-water LGE imaging~\cite{shaw2014left}, where joint fat-water separation and undersampled reconstruction pose additional challenges.